\documentclass[runningheads]{llncs}
\usepackage{geometry}
\usepackage{graphicx}
\usepackage{amsmath}
\usepackage{floatrow}
\usepackage[export]{adjustbox}
\usepackage[font=small]{subfig}
\usepackage{blindtext}

\begin{document}
\title{Object Contour and Edge Detection with RefineContourNet}
%
\author{Andr\'{e} Peter Kelm\inst{1}\orcidID{0000-0003-4146-7953} \and
Vijesh Soorya Rao\inst{2}\orcidID{0000-0003-4746-2694} \and
Udo Z\"olzer\inst{1} 
}
\authorrunning{Andr\'{e} Kelm et al.}
%
\institute{Helmut Schmidt University, Department of Signal Processing and Communications, Holstenhofweg 85, 22043 Hamburg, Germany \\
\email{\{andre.kelm, udo.zoelzer\}@hsu-hamburg.de}\\
\and
Hamburg University of Technology, Department of Mechatronics, \\ Am Schwarzenberg-Campus 1, 21073 Hamburg, Germany \\
\email{vijesh.rao@tuhh.de}}
\maketitle              
\begin{abstract}

A ResNet-based multi-path refinement CNN is used for object contour detection. For this task, we prioritise the effective utilization of the high-level abstraction capability of a ResNet, which leads to state-of-the-art results for edge detection. Keeping our focus in mind, we fuse the high, mid and low-level features in that specific order, which differs from many other approaches. It uses the tensor with the highest-levelled features as the starting point to combine it layer-by-layer with features of a  lower abstraction level until it reaches the lowest level. We train this network on a modified PASCAL VOC 2012 dataset for object contour detection and evaluate on a refined PASCAL-val dataset reaching an excellent performance and an Optimal Dataset Scale (ODS) of 0.752. Furthermore, by fine-training on the BSDS500 dataset we reach state-of-the-art results for edge-detection with an ODS of 0.824.

\end{abstract}
\section{Introduction}
Object contour detection extracts information about the object shape in images. Reliable detectors distinguish between desired object contours and edges from the background. Resulting object contour maps are very useful for supporting and/or improving various computer vision applications, like semantic segmentation \cite{SegmBoundaryFields,CED,DisFeatLearningVideoSegm}, object proposal \cite{MultiScaleGroupingObjProp} and object flow estimation \cite{OpticalFlow,CED}.

Holistically-Nested Edge Detection (HED) \cite{HED} has shown that it is beneficial to use features of a pre-trained classification network to capture desired image boundaries and suppressing undesired edges. Khoreva et al. \cite{WeaklySuperObjBound} have specifically trained the HED on object contour detection and proven the potential of HED for this task. Yang et al. have used a Fully Convolutional Encoder-Decoder Network (CEDN) to produce contour maps, in which the object contours of certain object classes are highlighted and other edges are suppressed more effectively than before \cite{CEDN}. Convolutional Oriented Boundaries (COB) \cite{COB} outperforms these results by using multi-scale oriented contours derived from a HED-like network architecture together with an efficient hierarchical image segmentation algorithm. A common feature in all this work is that a Very Deep Convolutional Network for Large-Scale Image Recognition (VGG) \cite{VGG} and its classifying ability is used as a backbone network. It is obvious that this backbone and its effective use are the major keys to the results achieved, but the new methods mentioned here do not use the latest classification networks, like Deep Residual Learning for Image Recognition (ResNet) \cite{ResNet}, which show a higher classification ability than VGG. We use a ResNet as backbone and propose a strategy to prioritise the effective utilization of the high-level abstraction capability for object contour detection. Accordingly we choose a fitting architecture and a customized training procedure. We outperform the methods mentioned previously and achieve a very robust detector with an excellent performance on the validation data of a refined PASCAL VOC \cite{pascal-voc-2012}. 
High-level edge detection is closely related to object contour detection, because object contours are often an important subset of the desired detection. Continuing, we will introduce the edge detection task and show that, unlike object contour detection, there is unexploited potential for using the high abstraction capability of classification networks.

Edge detection has a rich history and -- as one of the classic vision problems -- plays a role in almost any vision pipeline with applications like optical flow estimation \cite{BoundaryFlow}, image segmentation \cite{SemanticImSegmwithTaskSpecificEdgeDetection} or generative image inpainting \cite{EdgeConnect,EdgeGuided}. Classical low-level detectors, such as Canny \cite{canny1986computational} and Sobel \cite{sobel1970camera}, or the recently applied edge-detection with the Phase Stretch Transform \cite{PST}, filter the entire image and do not distinguish between semantic edges and the rest. Edge detection is no longer limited only to low-level computer vision problems. Even the evaluation method established to date -- the Berkeley Segmentation Dataset and the Benchmark 500 (BSDS500) \cite{gPb_early} -- requires high-level image processing algorithms for good results. Before Convolutional Neural Networks (CNNs) became popular, algorithms like the gPb \cite{gPb_early}, which uses contour detection together with hierarchical image segmentation, reached impressive results. In recent years, edge detectors, such as DeepNet \cite{kivinenDeepNet} and N\textsuperscript{4}-Fields \cite{Nfour} have begun to use operations from CNNs to reach a higher-level detection. DeepEdge \cite{DeepEdge} and DeepContour \cite{DeepContour} are CNN applications that use more high-level features to extract contours, and show that this capability improves the detection of certain edges. HED uses higher abstraction abilities than previous methods by combining multi-scale features and multi-level features extracted of a pre-learned CNN, and improves edge detection. Latest edge detectors such as the Crisp Edge Detector (CED) \cite{CED}, Richer Convolutional Features (RCF) \cite{RCF}, COB and a High-for-Low features algorithm (HFL) \cite{HFL} make use of their backbone classification nets for edge detection in different ways. But some of these networks are based on older backbone CNNs like the VGG and/or use simple HED-like skip-layer architectures. Similarly for object contour detection, we assume recent work has unexploited potential in the utilization of pre-trained classification abilities, in terms of architecture, backbone network, training procedure and datasets. We contribute a simple but decisive strategy, a network architecture choice following our strategy and unconventional training methods to reach state-of-the-art.
\\
Section 2 will briefly summarize the closest related work, Section 3 contains main contributions, like concept, realization and special training procedures for the proposed detector, Section 4 compares the method with other relevant methods and Section 5 concludes the paper.
\section{Related Work}

AlexNet \cite{AlexNet} was a breakthrough for image classification and was extended to solve other computer vision tasks, such as image segmentation, object contour, and edge detection. The step from image classification to image segmentation with the Fully Convolutional Network (FCN) \cite{FCN} has favored new edge detection algorithms such as HED, as it allows a pixel-wise classification of an image. HED has successfully used the logistic loss function for the edge or non-edge binary classification. Our approach uses the same loss function, but differs in term of another weighting factor, network architecture, and backbone network. Another image segmentation network, Learning Deconvolution Network for Semantic Segmentation \cite{Deconv}, has favored the development of the CEDN, demonstrating the strong relationship between image segmentation, object contour detection and edge detection. The good results of the CEDN inspired us to consider recent image segmentation networks for our task. Yang et al. created a new contour dataset using a Conditional-Random-Fields (CRF) \cite{CRF} refining method. CEDN and edge detector networks such as COB and HFL have an older backbone net and are outperformed by RCF, which is based on a ResNet and improved the edge detection. RCF has the same backbone network, but differs from our approach in using a different network architecture because it uses a skip-layer structure for feature concatenation like HED. We state that this simple concatenation is not effective enough for edge detection, and we propose to use a more advanced network structure. We have the hypothesis that an effective network architecture for edge detection should prioritise the high abstraction capability itself. As the deepest feature maps are the next ones to the classification layer, we propose to use them as the starting point to refine them layer-by-layer with features of a lower level until it reaches the level of classical edge detection algorithms. Our required properties are combined in RefineNet \cite{RefineNet} and that is why we have used the publicly available code from Guosheng Lin et al. as the basis of our approach. Parallelly to the implementation of our method, the CED from the work Learning to Predict Crisp Boundaries from Deng et al. \cite{LearnPCrispBound} has used a similar bottom-up architecture and surpasses RCF and achieves state-of-the-art. The work from Wang et al. Deep Crisp Boundaries \cite{CED} further develops this method and improved state-of-the-art results. Our approach mainly differs from theirs in its conceptualization. They also focus on producing "crisp" (thinned) boundaries, as they have shown that this benefits their results. We assume in contrast, that by focusing on the effective utilization of the high abstraction capability of a backbone network, we could achieve better results.

\section{RefineContourNet}
\subsection{Concept}
\begin{figure}[t]
\begin{center}
\includegraphics[width=0.8\textwidth]{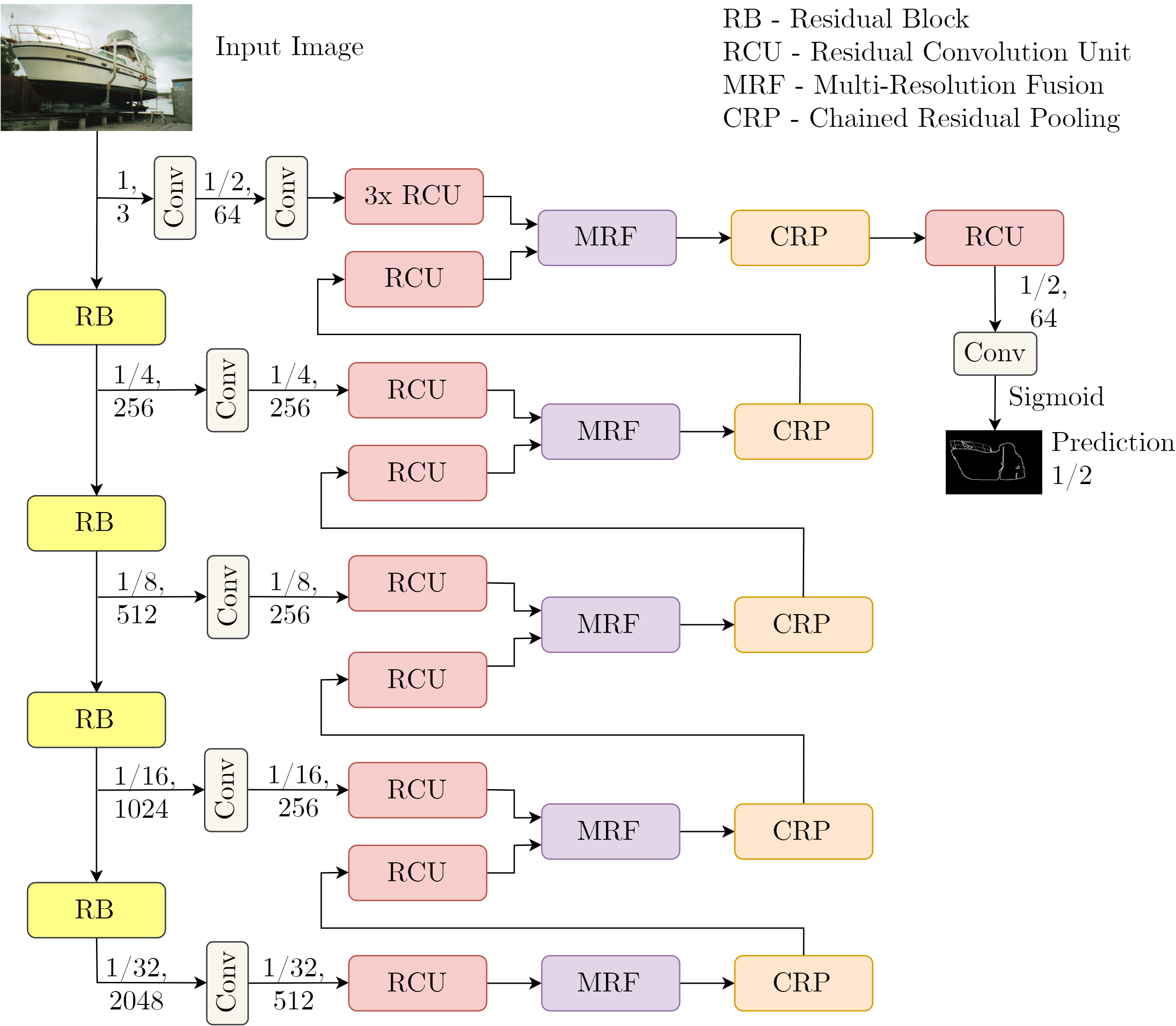}
\end{center}
\caption{RefineContourNet - modified RefineNet \cite{RefineNet}, where last layers are changed, such that there is one featue map at the end and a sigmoid activation function that predicts the probability of presence of contour} \label{fig1}
\end{figure}
Detecting edges with classical low-level methods can visualize the high amount of edges in many images. To distinguish between meaningful edges and undesired edges, a semantic context is required. Our selected contexts are the object contours of the 20 classes of the PASCAL VOC dataset. If context is clear and some low-level vision functions are available, the most important ability for an object contour detector is the high-level abstraction capability, so that edges can be distinguished in the sense of context. For this reason, our concept focuses on the effective use of the high-level abstraction ability of a modern classification network for object contour detection. With this strategy, we choose the architecture, backbone network, training procedure and datasets.

We hypothesize that an effective edge detection network architecture should prioritize the above mentioned capability. For this we propose to give preference to the deepest feature maps of the backbone network and to use them as the starting point in a refinement architecture. To connect the high-level classification ability with the pixel-wise detection stage, we assume, that a step-by-step refinement, where deep features are fused with features of the next shallower level until the shallowest level is reached, should be more effective than skip-layers with a simple feature concatination architecture. In most classification networks, features of different abstraction levels have different resolutions. To merge these features, a multi-resolution fusion is necessary. The RefineNet from Lin et al. \cite{RefineNet} provides the desired multi-path refinement and we base our application upon that and name our application in reference to this network RefineContourNet (RCN).

The training procedure has to accomplish two main goals: To effectively use the pre-trained features to form a specific abstraction capability for identifying desired object contours learned from data on the one hand and to connect this to the pixel-wise detection stage on the other hand. Because training data of object contours is limited, both training goals can be enhanced with data augmentation methods. For a similar reason, we do some experiments with a modified Microsoft Common Objects in Context (COCO) \cite{coco} dataset, to create an additional object contour dataset, usable for a pre-training. For fine-training on edge-detection, we offer a simple and unconventional training method that considers the individuality of BSDS500's hand-drawn labels.
\begin{figure}[]
\centering
\subfloat[RCU]{%
    \includegraphics[height=0.325\textwidth,valign=t]{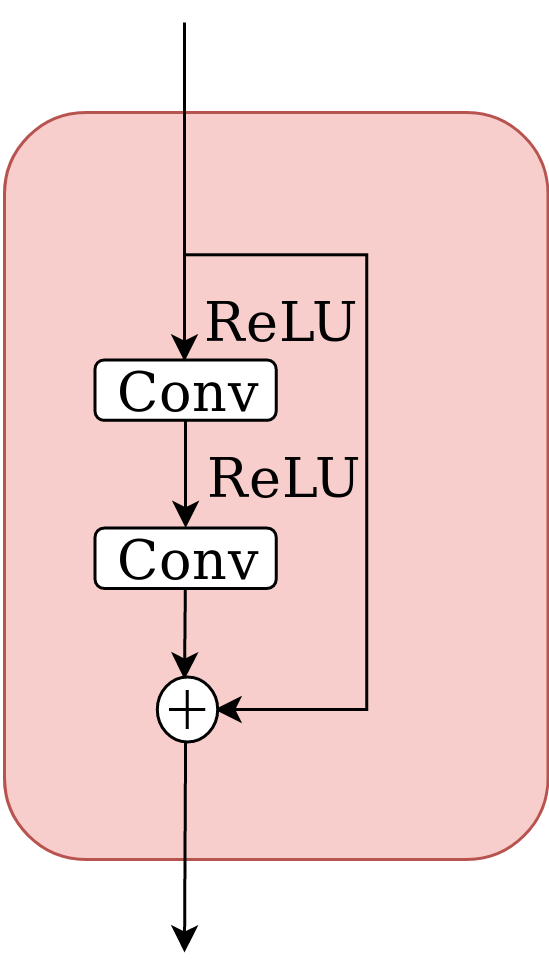}%
    \vphantom{\includegraphics[height=0.325\textwidth,valign=t]{figs/rcu.png}}%
  } \quad
  \subfloat[MRF]{%
    \includegraphics[height=0.325\textwidth,valign=t]{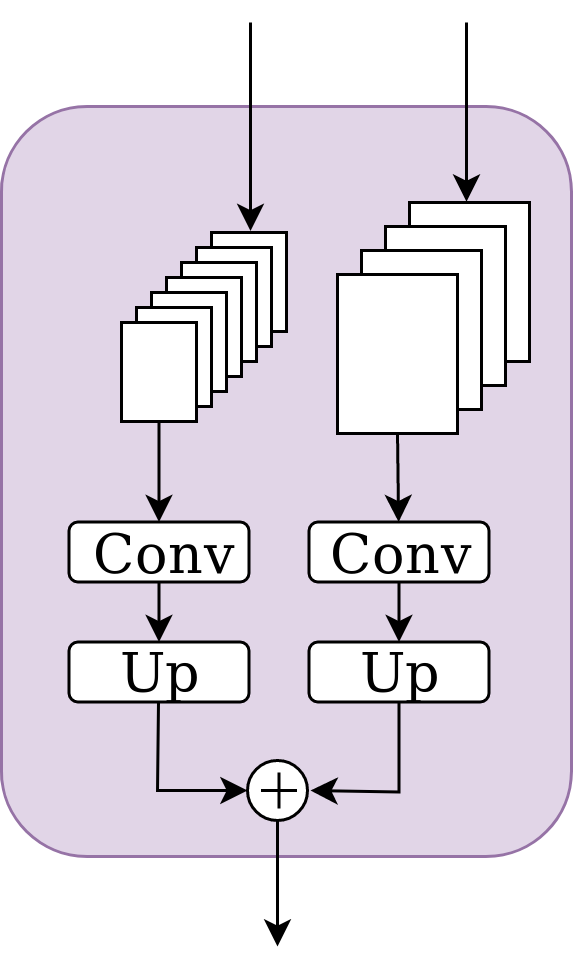}%
    \vphantom{\includegraphics[height=0.325\textwidth,valign=t]{figs/rcu.png}}%
  } \quad
  \subfloat[CRP]{%
    \includegraphics[height=0.325\textwidth,valign=t]{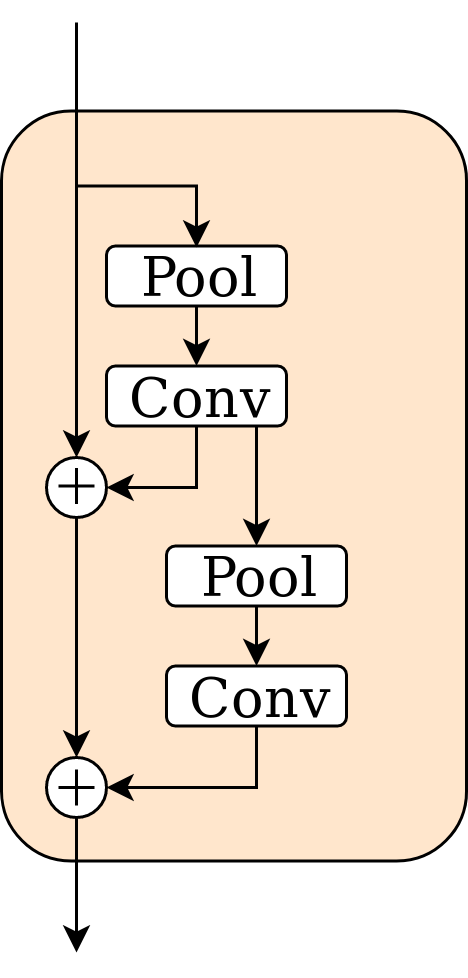}%
    \vphantom{\includegraphics[height=0.325\textwidth,valign=t]{figs/rcu.png}}%
  } \quad
  \caption{Block diagrams of refinement path operations\label{fig:blocks}}
\end{figure}
\subsection{Image Segmentation Network for Contour Detection}
The main difference between an image segmentation network and a contour detection network lies in the definition of the objective function. Instead of defining a multi-label segmentation, an object contour can be defined binary. We use the logistic regression loss function
\begin{align}
\begin{split}
		\textrm{L}(h_\Theta(x), y) & = -y \cdot \beta\log(h_\Theta(x)) \\
		& \quad - (1-y) \cdot \log(1-h_\Theta(x)),
			\end{split} 
			\label{eq:LL2}  
\end{align}
with $h_\Theta(x) \in [ \, 0, 1]$, $y \in \{ \, 0, 1\}$ and $\beta = 10$, where $h_\Theta(x)$ is the prediction for a pixel $x$ with the corresponding binary label $y$.  $\Theta$ symbolizes the learned parameters and $\beta$ is a weighting factor for enhancing the contour detection due to the large imbalance between the contour and the non-contour pixels. Changing the loss function results in a change of the last layer of the RefineNet according to the binary loss function. The 21 feature map layers previously used to segment 20 PASCAL-VOC classes, including background class, will be replaced by a single feature map sufficient for binary classification of contours.
\subsection{Network Architecture}
Figure \ref{fig1} shows the RefineContourNet. For clarity, the connections between the blocks specifies the resolution of the feature maps and the size of the feature channel dimension. The Residual Blocks (RB) are part of the ResNet-101. RefineNet has introduced three different refinement path blocks: The Residual Convolution Unit (RCU), the Multi-Resolution Fusion (MRF) and the Chained Residual Pooling (CRP). They are arranged in a row to use the higher-level features as input to combine them with the lower-level features of the RB at the same level. The RCU in Fig. \ref{fig:blocks} (a) has residual convolutional layers and enriches the network with more parameters. It can adjust and modify the input for MRF. The MRF block adapts the input first by performing a convolution operation, in order to adjust the channel dimension of the feature space corresponding to the higher-level ones with the lower level. Then, the smaller resolution feature maps get upsampled to have same tensor dimensions as the larger ones, after which they are added, as shown in Fig. \ref{fig:blocks} (b). The goal of the CRP is to gather more context from the feature maps than a normal max pooling layer. Several pooling blocks are concatenated and each block consists of a max-pooling with higher stride length and a convolutional operation. Illustration of CRP with two max-pooling operations is shown in Fig. \ref{fig:blocks} (c). In the final refinement step, we use the original image as input for an extra path with 3 RCUs, which improves the results.
\section{Evaluation}
\begin{figure}[t]
\centering
\subfloat[Original]{%
    \includegraphics[height=0.51\textwidth,valign=t]{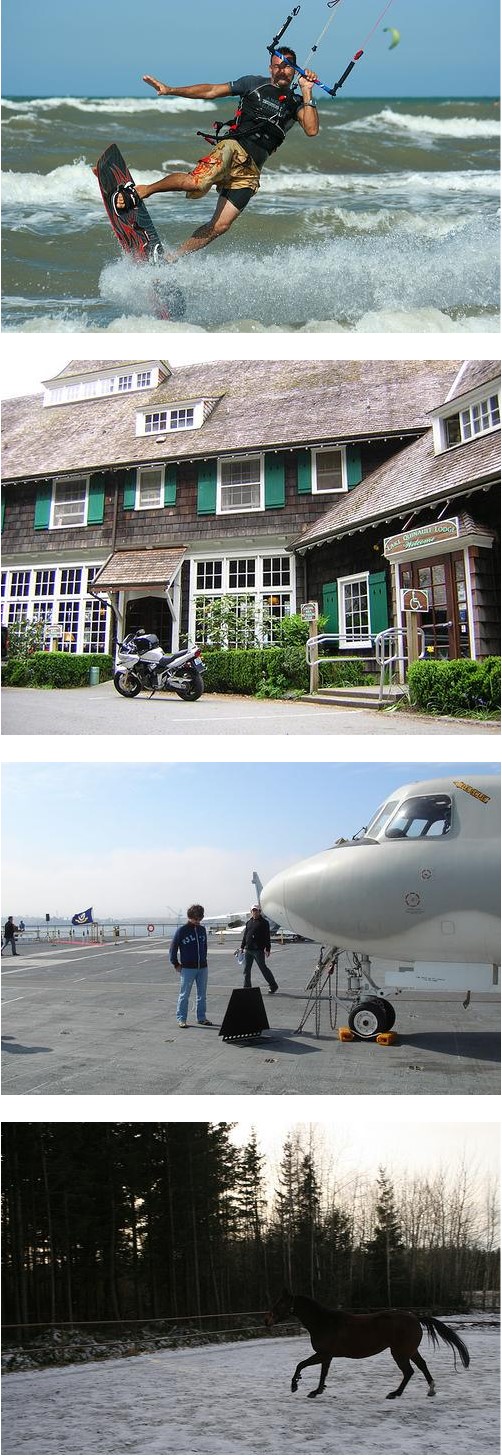}%
    \vphantom{\includegraphics[height=0.5\textwidth,valign=t]{figs/PASCALORIGCAIP.jpg}}%
  } \quad
  \subfloat[GT]{%
    \includegraphics[height=0.51\textwidth,valign=t]{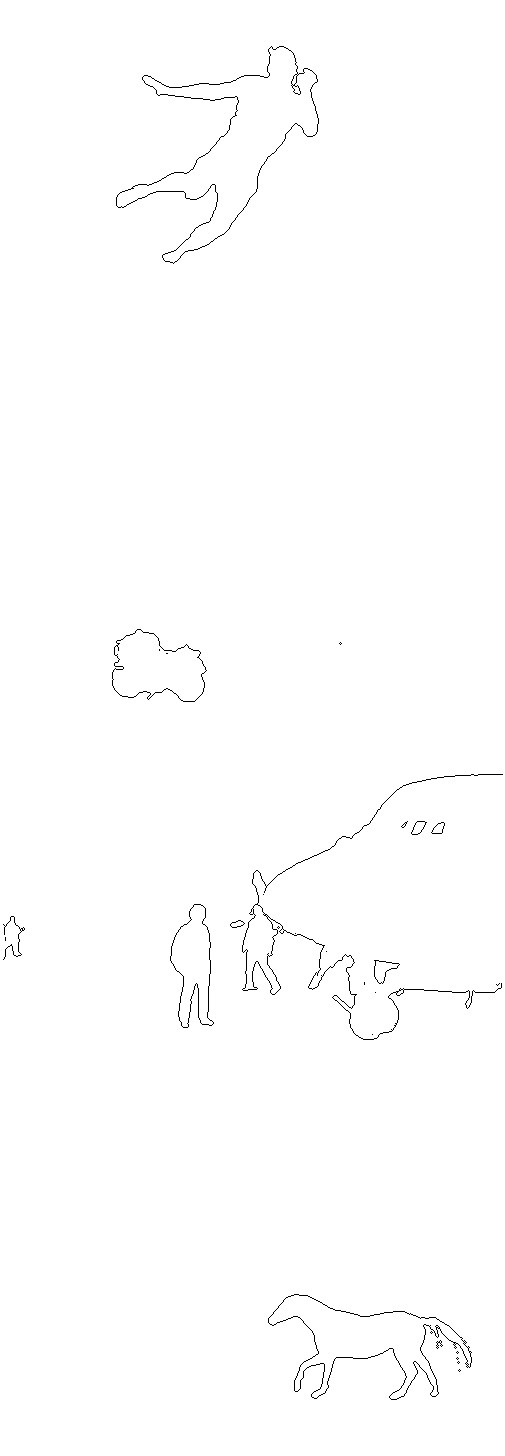}%
    \vphantom{\includegraphics[height=0.5\textwidth,valign=t]{figs/PASCALORIGCAIP.jpg}}%
  } \quad
  \subfloat[CEDN]{%
    \includegraphics[height=0.51\textwidth,valign=t]{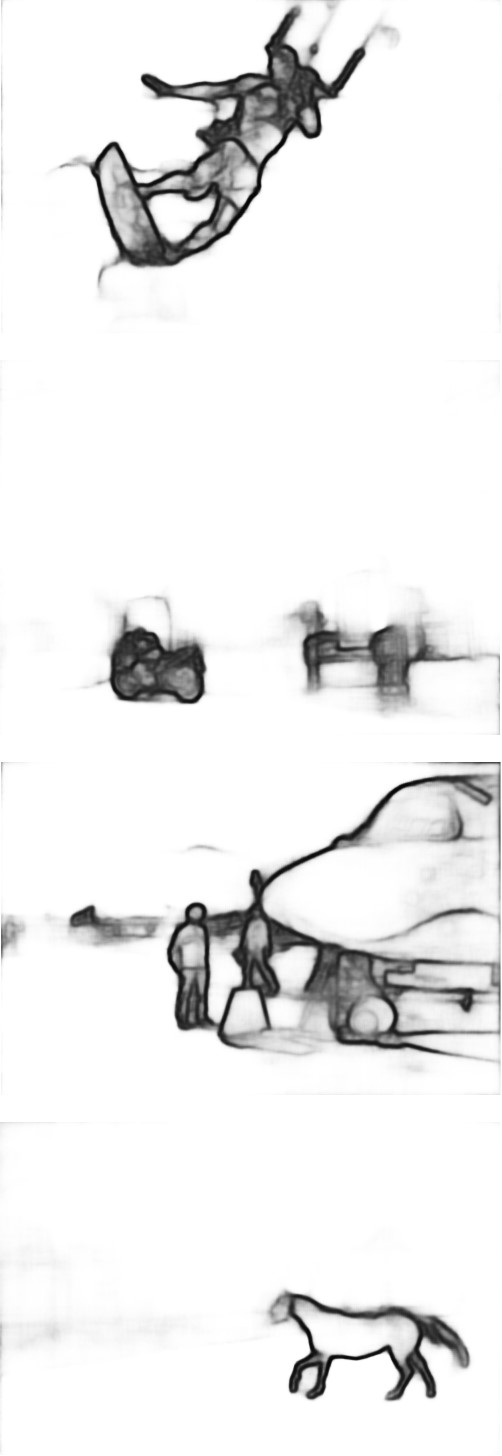}%
    \vphantom{\includegraphics[height=0.5\textwidth,valign=t]{figs/PASCALORIGCAIP.jpg}}%
  } \quad
    \subfloat[RCN-VOC]{%
    \includegraphics[height=0.51\textwidth,valign=t]{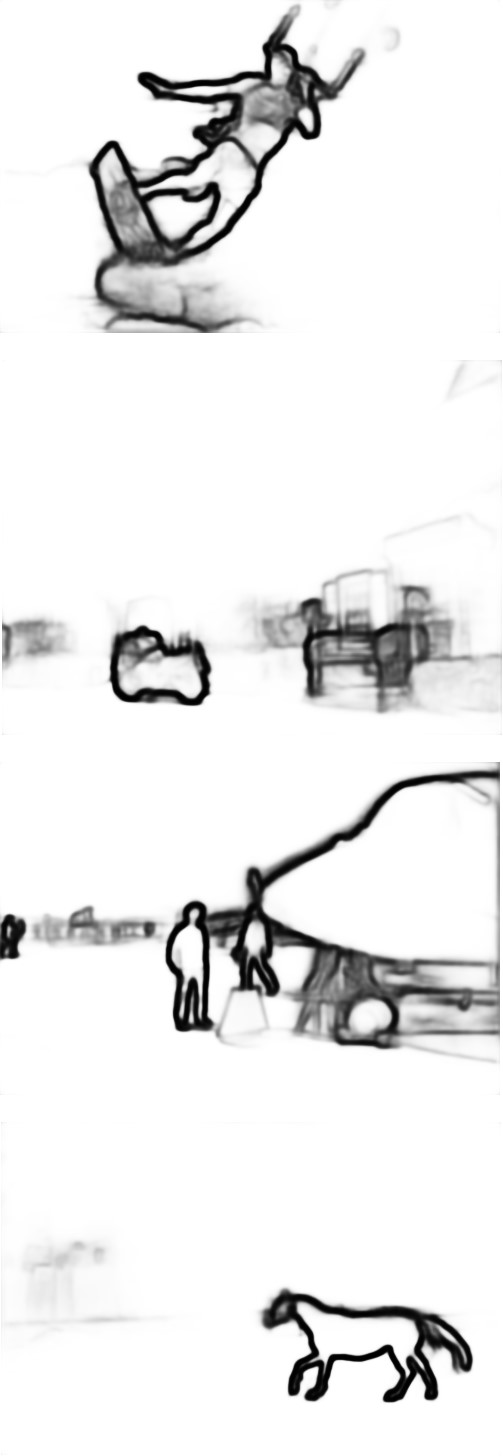}%
    \vphantom{\includegraphics[height=0.5\textwidth,valign=t]{figs/PASCALORIGCAIP.jpg}}%
  } \quad
    \subfloat[RCN]{%
    \includegraphics[height=0.51\textwidth,valign=t]{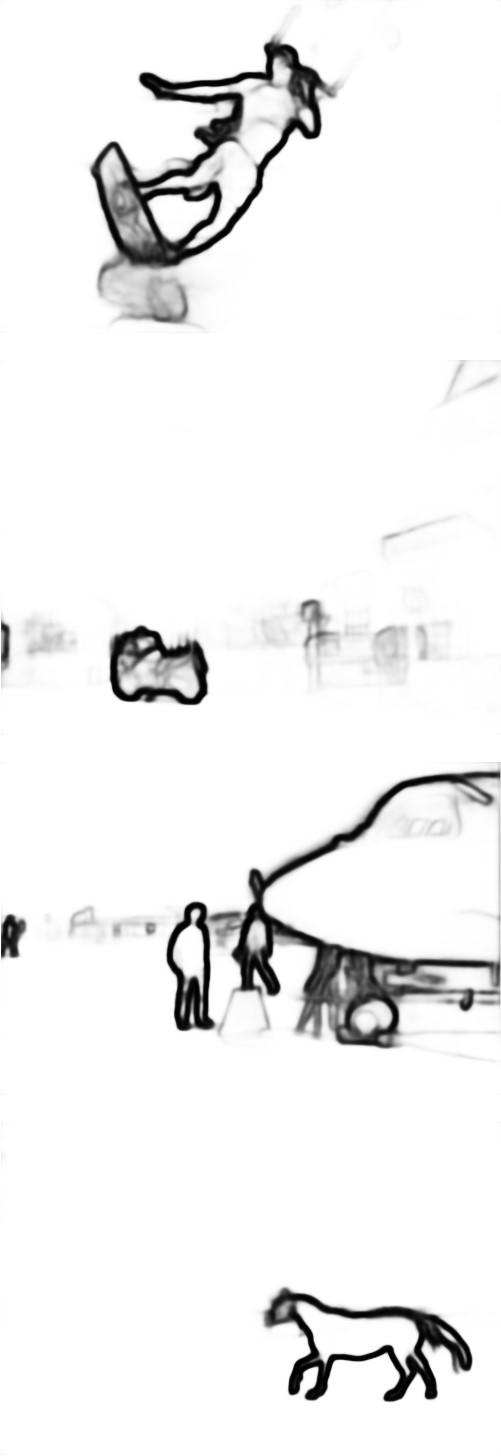}%
    \vphantom{\includegraphics[height=0.5\textwidth,valign=t]{figs/PASCALORIGCAIP.jpg}}%
  } \quad
  \caption{Visualization of object contour detection methods\label{fig:objectcont}}
\end{figure}
We have done various experiments with different combinations of refinement blocks per multipath, and we have always observed the best results by placing the three blocks sequentially in a row, as shown in Fig. \ref{fig1}. 
\subsection{Training}
For each epoch, 1000 random images are selected from the training set, and the data is augmented by random cropping, vertical flipping, and scaling between 0.7 and 1.3. To find an optimal training method, we have examined the following training variants:
\begin{itemize}
\item \textbf{RCN-VOC} is trained only on the CRF-refined object contour dataset proposed by Yang et al. \cite{CEDN}.
\item \textbf{RCN-COCO} is pre-trained on a modified COCO dataset, where we have considered only the 20 PASCAL VOC classes and have produced contours. COCO segmentation masks and from those generated contours are not accurate, so we enrich them with additional contours. For this we use our own object contour detector RCN-VOC and set a high threshold to add only confident contour detections.
\item \textbf{RCN} is pre-trained on the modified COCO and trained on the refined PASCAL VOC.
\end{itemize}

Training the network for edge detection involves fine-training on the validation and train sets of the BSDS500 dataset. The BSDS contains individual, hand-drawn contours for the same images created by different people. We take the subjective decisions into account of which edge is a desired edge and which is not, by simply using all individual labels, and let the CNN form a compromise. To give an indication of how such training affects the results, we fine-train one of the networks only on the drawings of one single person, called \textbf{RCN-VOC-1}. All trainings and modifications are done in MatConvNet \cite{vedaldi15matconvnet}.
\subsection{Object Contour Detection Evaluation}
\begin{figure}[t]
\CenterFloatBoxes
\begin{floatrow}
\ffigbox[\FBwidth]
 {\includegraphics[trim=10 2 10 4,clip,width=0.6\textwidth]{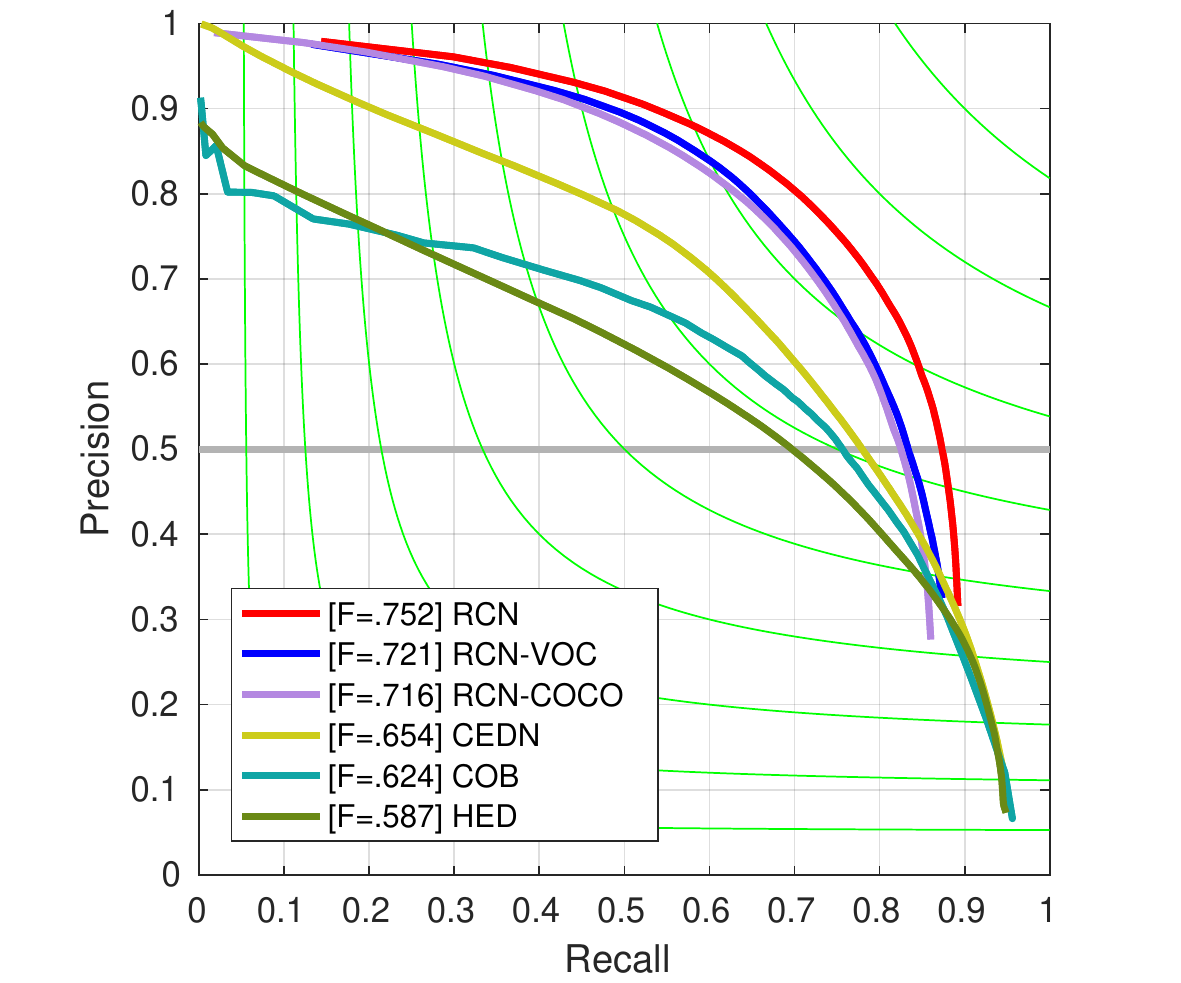}}
  {   \caption{PR-curves on refined PASCAL val2012}\label{fig:PRPascal}}
\killfloatstyle
\ttabbox
  {\begin{tabular}{|l|c|c|c|}
\hline
Net & ODS & OIS & AP \\
\hline\hline
          \textbf{RCN} & \textbf{.752} & \textbf{.773} & .641\\
          RCN-VOC & .721 & .746 & .613 \\
          RCN-COCO & .716 & .741 & \textbf{.719}\\
          CEDN & .654 & .657 & .679\\
          COB & .624 & .657 & .593\\
          HED & .587 & .598 & .568\\
\hline
\end{tabular}
  }
  {\caption{Comparison on refined PASCAL val2012}\label{tab:Pascaltab}}
\end{floatrow}
\end{figure}
For evaluation, we use Piotr's Computer Vision Matlab Toolbox \cite{Dollar}, the included Non-Maximum-Suppression (NMS) algorithm for thinning the soft object contour maps and a subset of 1103 images of a CRF-refined PASCAL val2012. We calculate the Precision and Recall (PR) curve for the RCN models, CEDN, HED and COB in Fig.\ref{fig:PRPascal}. In Tab. \ref{tab:Pascaltab} the Optimal Dataset Scale (ODS), Optimal Image Scale (OIS) and the Average Precision (AP) for the methods are noted. The quantitative analysis reveals that the RCN models significantly perform better in comparison to the other methods on all three metrics. This is also reflected in the visual results, cf. Fig. \ref{fig:objectcont}. The RCN-VOC and RCN have upper hand in suppressing the undesired edges, such as inner contours of the objects. At the same time, they also can recognize object contours more clearly. A disadvantage is that the contour predictions are thicker than in CEDN, which is due to the halved resolution owed by the network architecture. Nevertheless, the detection is very robust and the NMS can effectively calculate 1-pixel thinned object contours.

\subsection{Edge Detection Evaluation}
\begin{figure}[]
\centering
\subfloat[Original]{%
    \includegraphics[height=1.03\textwidth,valign=t]{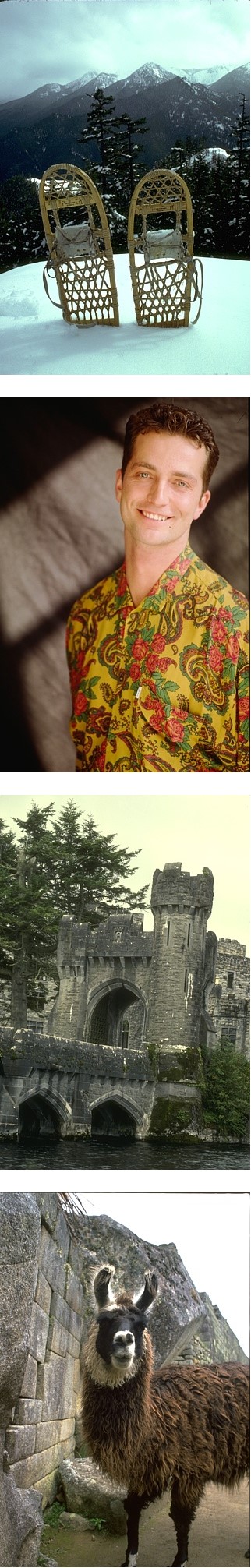}%
    \vphantom{\includegraphics[height=1\textwidth,valign=t]{figs/BSDS_CAIP_ORIG.jpg}}%
  } \quad
  \subfloat[GT]{%
    \includegraphics[height=1.03\textwidth,valign=t]{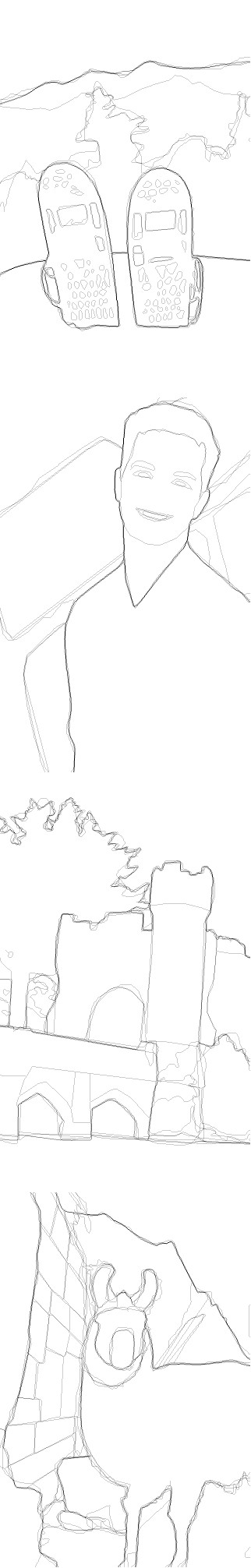}%
    \vphantom{\includegraphics[height=1\textwidth,valign=t]{figs/BSDS_CAIP_ORIG.jpg}}%
  } \quad
  \subfloat[CED]{%
    \includegraphics[height=1.03\textwidth,valign=t]{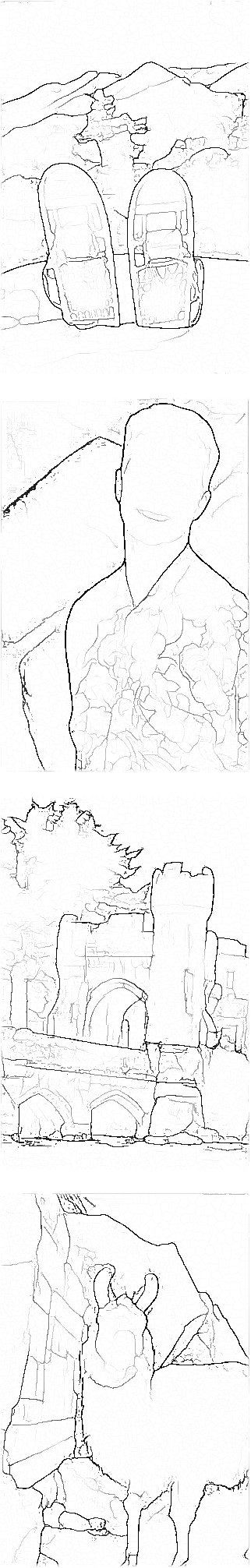}%
    \vphantom{\includegraphics[height=1\textwidth,valign=t]{figs/BSDS_CAIP_ORIG.jpg}}%
  } \quad
    \subfloat[RCN-VOC]{%
    \includegraphics[height=1.03\textwidth,valign=t]{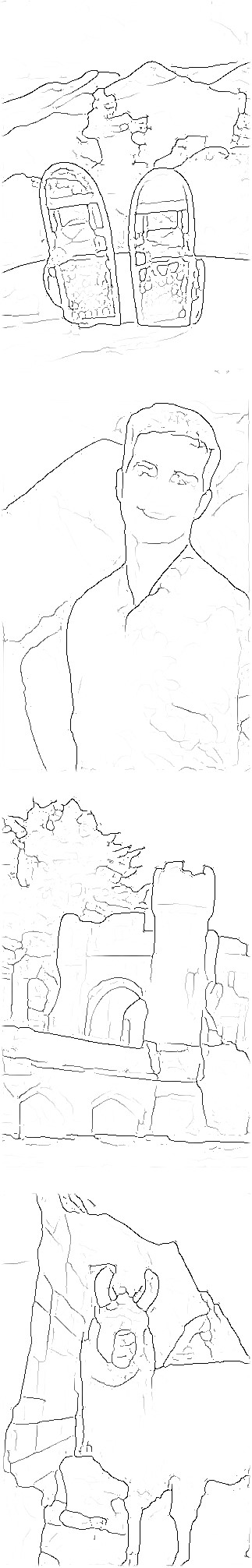}%
    \vphantom{\includegraphics[height=1\textwidth,valign=t]{figs/BSDS_CAIP_ORIG.jpg}}%
  } \quad
  \caption{Visualization of edge detection methods\label{fig:visualedgedet}}
\end{figure}

\begin{figure}[t]
\CenterFloatBoxes
\begin{floatrow}
\ffigbox[\FBwidth]
 {\includegraphics[trim=10 2 10 4,clip,width=0.6\textwidth]{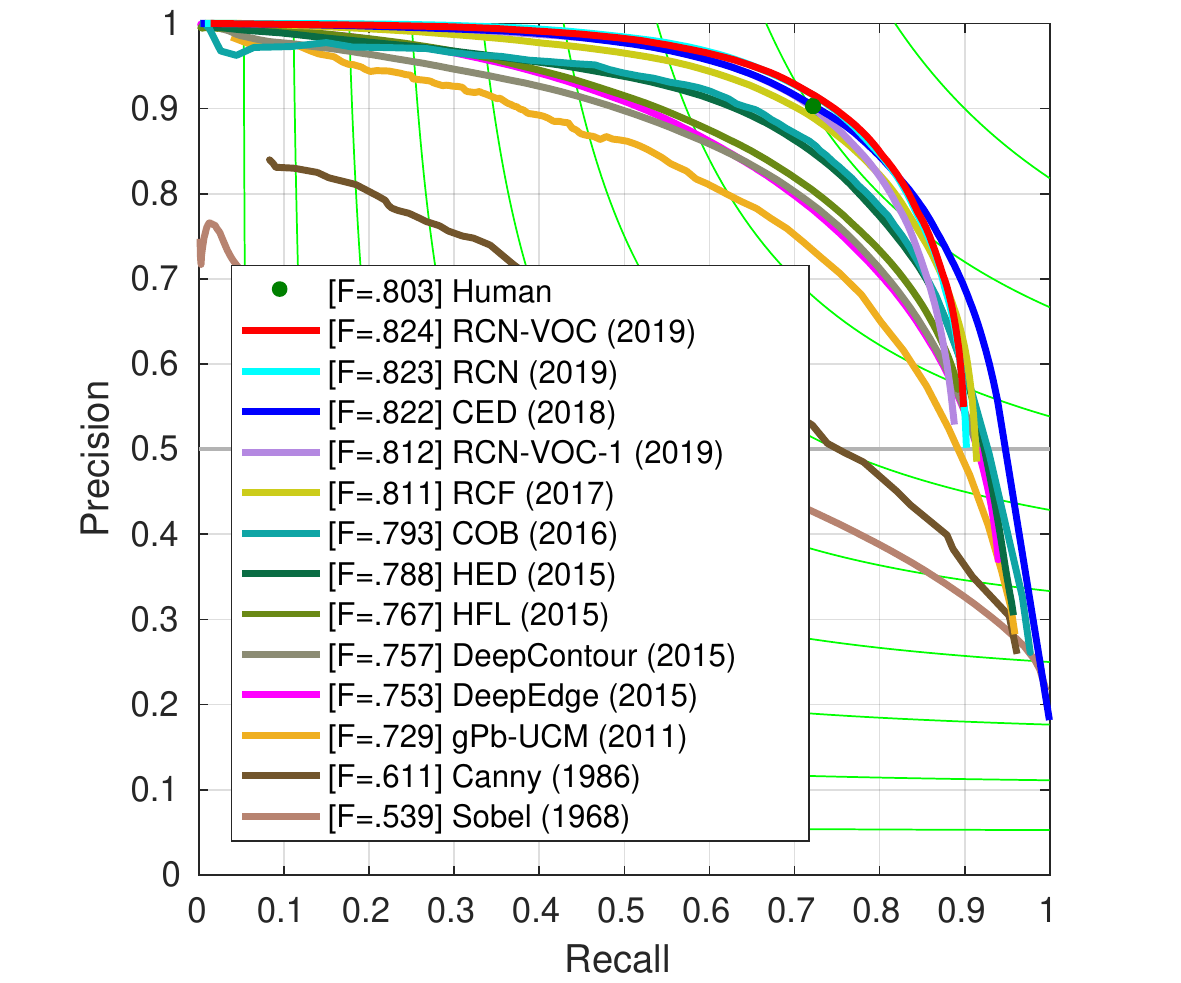}}
  {   \caption{PR-curves on BSDS500}\label{fig:PRBSDS}}
\killfloatstyle
\ttabbox[\Xhsize]
  {\begin{tabular}{|l|c|c|c|}
\hline
Net & ODS & OIS & AP \\
\hline\hline
          \textbf{RCN-VOC} & \textbf{.824} & .839 & .837\\
          RCN & .823 & .838 & .853\\
          CED & .822 & \textbf{.840} & \textbf{.895}\\
          RCN-VOC-1 & .812 & .827 & .822\\
          RCF & .811 & .830 & .846 \\
          Human & .803 & .803 & - \\
          COB & .793 & .819 & .849\\
          HED & .788 & .808 & .840\\

\hline
\end{tabular}
  }
  {\caption{Comparison on BSDS500}\label{tab:BSDS}}
\end{floatrow}
\end{figure}

The results of a quantitative evaluation of the RCN on unseen test images of BSDS500 are represented in the PR-curves in Fig \ref{fig:PRBSDS}. ODS, OIS and AP from methods such as RCN, CED, RCF, COB and HED are listed in Tab. \ref{tab:BSDS}. The proposed RCN achieves the state-of-the-art with a higher ODS than recent methods, closely followed by CED. In Fig. \ref{fig:visualedgedet} results of CED and RCN are visualized for some test images. Careful analysis of the results reveals that RCN detects some relevant edges, cf. inner contours of the snowshoes (1st row), face of the young man (2nd row), snout of the llama (4th row), which CED no longer recognizes. As for the object contour detection task, the disadvantage of the thicker edge predictions persists for the RCN. However, the NMS works more precisely for edge prediction maps from RCN, as the bit depth per pixel is increased from 8 to 16 bits. The difference is evident in the results for background edges in the image of the young man (2nd row), since an absolute maximum of the CED prediction could not be clearly distinguished, RCN edges are thinned more effectively.

\section{Conclusion}
The strategy, of using the high abstraction capability for object contour and edge detection more effectively than previous methods, has given us very good results in object contour detection and state-of-the-art results in edge detection. Our concept that RefineNet \cite{RefineNet} provides a very useful  bottom-up multipath refinement architecture for edge detection is supported by these results. With the unconventional training methods, like the pre-training with a modified COCO dataset or by simply using all individual labels for fine-training on BSDS500, we have been able to improve the respective task.
%
%
%
\clearpage
\bibliographystyle{splncs04}
\bibliography{egbib}
%
%
%
%
\end{document}